\documentclass{article}

\usepackage[preprint]{nips_2018}

\usepackage{url}            
\usepackage{booktabs}       
\usepackage{amsfonts}       
\usepackage{nicefrac}       
\usepackage{microtype}      

\usepackage{multibib}

\usepackage{times}
\usepackage{graphicx} 
\usepackage{subfigure} 

\usepackage{enumitem}

\usepackage{makeidx}
\usepackage{lipsum, amsmath,amssymb} 
\usepackage{color}

\usepackage{epsfig}

\usepackage{mathtools}

\usepackage{relsize}
\usepackage{multirow}
\usepackage{bm}

\usepackage{stmaryrd}
\usepackage{algorithm}
\usepackage{algorithmic}
\newenvironment{myitemize}
{   \begin{itemize}[noitemsep,topsep=0pt]
	\setlength{\itemsep}{0pt plus 0pt} 
    \setlength{\parskip}{0pt}
    \setlength{\parsep}{0pt}     }
{ \end{itemize}                  }

\usepackage{hyperref}

\newcommand{\icol}[1]{
  \begin{smallmatrix}#1\end{smallmatrix}%
}

\title{Network Learning with Local Propagation}

\author{
  Dimche Kostadinov,  
   Behrooz Razeghi, Sohrab Ferdowsi, Slava Voloshynovskiy    \\
  Department of Computer Science,  University of Geneva, 
  Geneva, Switzerland \\ 
  {e-mail}: \texttt{$\{$Dimche.Kostadinov, Behrooz.Razeghi,}\\
  \texttt{ Sohrab.Ferdowsi, svolos$\}$@unige.ch} 
}

\begin{document}

\maketitle

\begin{abstract}


This paper presents a locally decoupled network parameter learning with local propagation. Three elements are taken into account: (i) sets of nonlinear transforms that describe the representations at all nodes, (ii) a local objective at each node related to the corresponding local representation goal, and (iii) a local propagation model that relates the nonlinear error vectors at each node with the goal error vectors from the directly connected nodes. The modeling concepts (i), (ii) and (iii) offer several advantages, including: (a) a unified learning principle for any network that is represented as a  graph,  (b) understanding and interpretation of the local and the global learning dynamics, (c) decoupled and parallel parameter learning, (d) a possibility for learning in infinitely long, multi-path and multi-goal networks. Numerical experiments validate the potential of the learning principle. The preliminary results show advantages in comparison to the state-of-the-art methods, w.r.t. the learning time and the network size while having comparable recognition accuracy.
\end{abstract}

\vspace{-.1in}
\section{Introduction}\label{sec:Introduction}
\vspace{-.1in}


In the recent years, the multi-layer neural networks have had significant progress and advances, where impressive results were demonstrated on variety of tasks across many fields \citet{Schmidhuber:2014:Overview}.  
A multi-layer neural network has a target that is defined by a loss function which most often is specified in a supervised manner, and is set for the representation at the last node in the network.
As a general practice, back-propagation \citet{Plaut:1986:BP}, \citet{Lecun::TB:BP} and \citet{Schmidhuber:2014:Overview} is applied to learn the parameters of the network. Commonly, a gradient-based algorithm \citet{LeCun:1998:EBT}, \citet{Yoshua:2012:PR:GBT} is used to optimize a non-convex objective. That is, the gradient of 
the loss is sequentially propagated from the last node throughout the network nodes back to the first node. 

One of the most crucial issues in back-propagation is the vanishing gradient \citet{Hochreiter:1998:VGP}  and the exploding  gradient \citet{Razvan:Pascanu:Exp:Grad:prob} that might lead to a non-desirable local minima (or saddle point).
On the other hand, the dependencies from the subsequent propagation make this approach not suitable for parallel parameter learning per node.  An additional challenge is the interpretation of the dynamics during training. 
Several works \citet{Bottou:2012:SGD_Tricks}, \citet{Shamir:2013:SGD_NON_Smood}, \citet{Srivastava:2014:Dropout},\citet{Kingma:2014:Adam}, \citet{Loshchilov:2016:SGDR}, \citet{Ruder:2016:Overview} \citet{Goh:2017:why_MW},\citet{Zhu:Adam_Look_Ahead} 
have addressed these issues and proposed improvements. 
However, they all are within the realm of the concept that is defined by a goal (target) at the last node in the network (we point out to \citet{Schmidhuber:2014:Overview} for an 
overview). 
Even the concept of the recent works by \citet{JaderbergCOVGK:2016} and \citet{CzarneckiSJOVK:2017:Sintetic:Gradient} 
falls in this category, 
together with the methods proposed by \citet{LeeZ:2014TP}, \citet{Balduzzi:2015:KCB}, \citet{Taylor:2016BXSPG} and \citet{Nokland:2016DF}. 

\begin{figure}[t]
\centering
\begin{center}
\vskip -0.08in
\begin{minipage}[b]{0.52\linewidth}
\centering
\centerline{\includegraphics[width=\columnwidth]{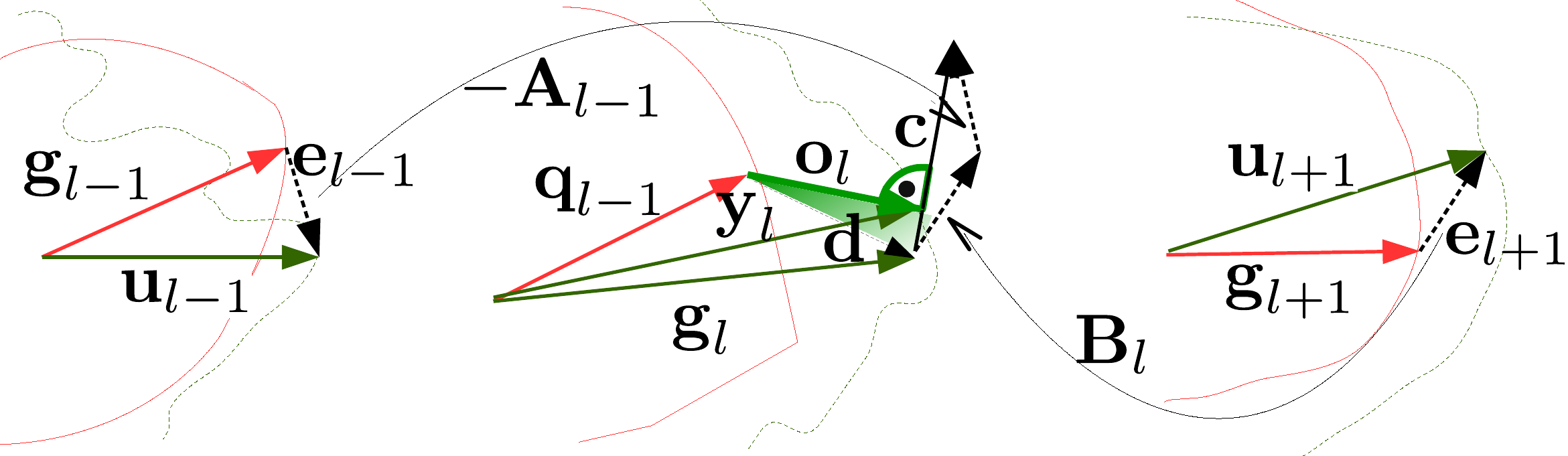}}
\vskip -0.0in
\begin{tabular}{c@{\hspace{60pt}}c@{\hspace{60pt}}c}
$l-1$ & $l$& $l+1$ \\
& a)&
\end{tabular}
\end{minipage}
\label{prior:dinimics:ilustration}
\begin{minipage}[b]{0.49\linewidth}
\centering
\end{minipage}
\begin{minipage}[b]{0.46\linewidth}
\centering
\centerline{\includegraphics[width=\columnwidth]{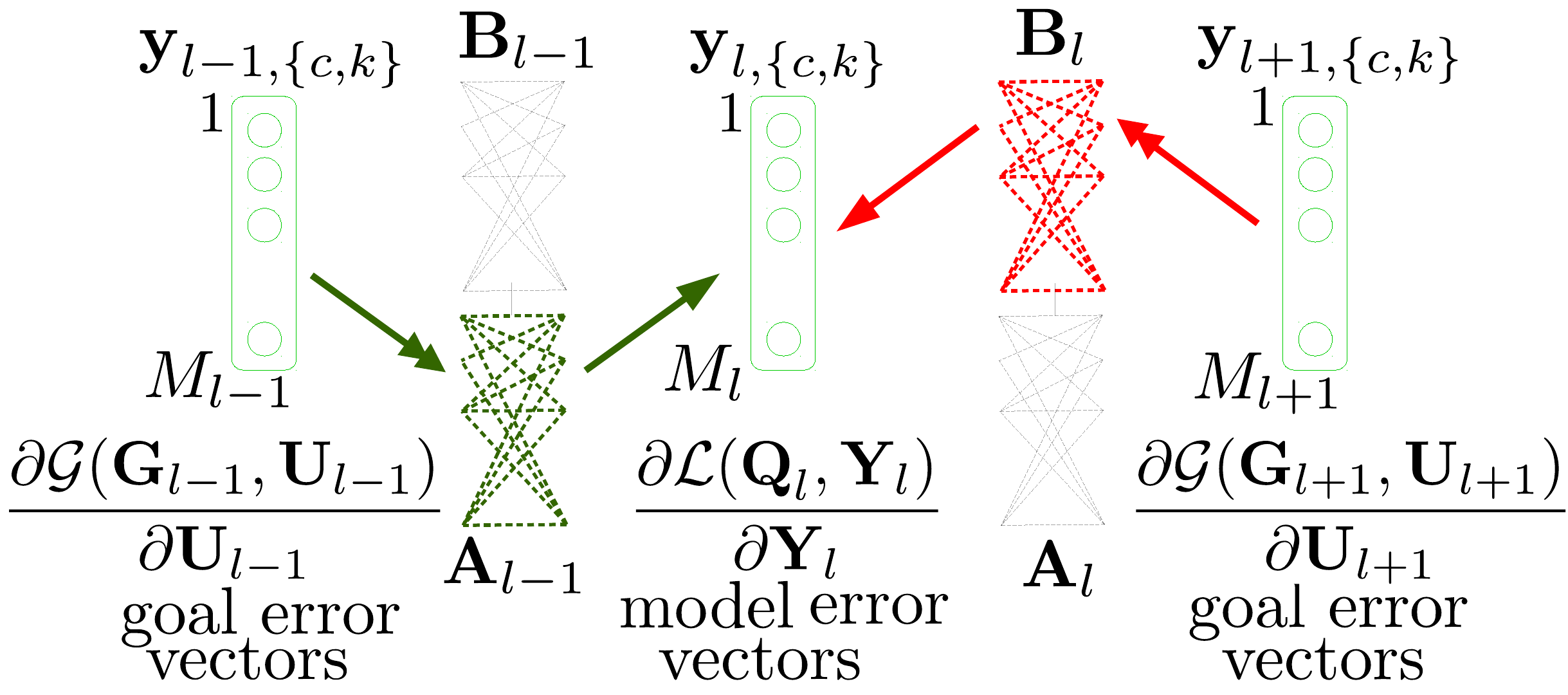}}
b)
\end{minipage}
\label{intro:image:prior:dynamics}
\end{center}
\vskip -0.19in
\caption{a) 
The red and green curves represent the space were the nonlinear transform and the desired representations live, respectively. The goal error vectors at node levels $l-1$ and $l+1$ are 
${\bf e}_{l-1}={\bf u}_{l-1}-{\bf g}_{l-1}$
 and 
${\bf e}_{l+1}={\bf u}_{l+1}-{\bf g}_{l+1}$. The change of the local propagation flow is denoted as ${\bf c} \backsimeq {\bf B}_{l}{\bf e}_{l+1}+{\bf A}_{l-1}{\bf e}_{l-1}$, the transform error vector is ${{\bf o}_l={{\bf A}_{l-1}{\bf u}_{l-1}-{\bf y}_{l}}}$ and ${\bf c}^T{\bf o}_l$ 
is the local propagation term \eqref{eq:global:problem:error:propagation:flow}.
The set of directions colored in green highlights a trade-off. b) The resulting 
local network when fixing all the network variables except ${\bf A}_{l-1}, {\bf B}_{l}$ and ${\bf Y}_l$.  The 
direction of the forward and backward propagation flow is marked with green and red, ${\bf U}_{{l-1}}$ and ${\bf U}_{{l+1}}$ are the sparse representations.
}
\vskip -0.2in
\end{figure}

\textbf{Addressed Question}  In this paper, we ask the question $-$\textit{
What are the essential elements and fundamental trade-offs to a learning principle that also uncover its learning dynamics, in local or global context w.r.t. representations at the network nodes?
}  
$-$ 
towards answering it, in the following, we introduce several concepts, present novel learning principle, give new results and present numerical evaluation.

\vspace{-0.15in}
\subsection{Network Modeling and Learning Approach Overview} 
\vspace{-0.07in}
We consider a feed-forward network consisting of nodes and weights as connections between the nodes that has two operational modes learning: ($Lm$)  and testing ($Tm$). 

\vspace{-0.15in}
{\flushleft \textbf{Representation Description Per Network Node}}: 
At $Lm$, the description of each representation at each network node cores on three elements 
(i) \textit{generalized nonlinear transforms} (gNTs),  sparsifying nonlinear transforms (sNTs) \citet{RubinsteinE14} and \citet{DBLP:conf/icassp/RavishankarB14} 
(ii) \textit{local goal}  (iii) and \textit{local propagation component}. At ($Tm$) we use only sNT. 

\vspace{-0.07in}
{\textit{$-$ Generalized Nonlinear Transform}}: We introduce gNT to represents a principal description of an element-wise nonlinearity that is 
analogous to the commonly used description by an activation function. 
The advantage is that a gNT offers a 
high degree of 
freedom in modeling\footnote{Many nonlinearities,  \textit{i.e.}, ReLu, ${\ell_p}$-norms, elastic net-like, $\frac{\ell_1}{\ell_2}$-norm ratio, binary encoding, ternary encoding, etc., can be modeled as a generalized nonlinear transform representation.} and imposing constraints. 
At the same time it allows the used constraints, if any, to be interpreted, explained 
and connected to 
an empirical risk \citet{Vapnik:1995:SL:ER}. 

\vspace{-0.05in}
\textit{$-$ Local Goal}: 
The \textit{local goal} per representation 
describes the 
desired representation per node 
that is formally defined w.r.t. a linear transform representation at that node and 
a function. A key here is that we use a function analogous to the concept of objective, but, the difference is that we define the functional mapping as a solution to an optimization problem, where its role is to transform a given representation into a representation with specific properties (\textit{e.g.} discrimination, information preserving, local propagation constraints preserving, sparsity, compactness, robustness etc.). 
The measured difference between the desired representation (or gNT representation) and the sNT representation (or the linear transform representation) 
specifies the \textit{local goal error} (or \textit{nonlinear transform error}).
At each node it is possible, but, not necessary, to define a goal for the corresponding representation. 


\vspace{-0.06in}

\textit{$-$ Local Propagation}:
The \text{local propagation} is modeled by the inner product between the transform error vectors at each node and the affine combination of the propagated goal error vectors from the closely connected nodes. 
The motivation is to allow 
an independent and 
decoupled estimation of the parameters per 
the local networks consisting of a node and its directly connecting nodes. 

\vspace{-0.04in} 
Computational complexity-wise the gNT representation estimation with (or without) a local goal
and a local propagation constraints is a low complexity constrained projection problem. 
\vspace{-0.04in}
{ \textbf{Learning with Nonlinear Transforms and Local Propagation}  }
We present a novel general learning problem formulation for estimation of the network parameters that includes, both the gNTs and the sNTs. The learning target is to estimate the network parameters such that the sets of gNTs reduce to the a set of sNTs and all the local representation goals are achieved. 
In fact, we show that the expected mismatch between the sets of gNTs and the set of sNT representations, that in fact is addressed by the learning problem, can be seen as an empirical risk for the network during $Tm$. 

\vspace{-0.07in}
{ \textit{$-$ Locally Decoupled Parallel Learning Strategy}}:
We propose a novel learning strategy consisting of two stages:  (i) estimation of sNT representations and (ii)  estimation of the parameters in the gNTs with the possibility for operation in synchronous and asynchronous mode. At the first stage, the solution is a simple propagation through the network by successively using the corresponding sNT per each node. At the second stage, we propose a solution 
that involves 
local estimation per subsets of the network parameters. One common decoupled problem is addressed  with local propagation constraints and 
solved by an iterative, alternating algorithm with three steps. We propose exact and approximate closed form solutions for the respective steps. 

\vspace{-0.07in}
\textit{$-$ Interpretations and Learning Dynamics}: The local propagation is explained by showing a connection to a \textit{local diffusion model} \citet{Kittel:Charles_Kroemer:Herbert_1980} or change of the local flow. 
At each node, it reflects the change of the desired properties of the representations, which is formally defined by a goal, within a node and its neighboring nodes. 

\vspace{-0.07in}
{ \textit{$-$ Evaluation of the Learning Principle}}: We theoretically and empirically validate that local learning with a proper constraint on the local propagation can be used to achieve desirable global data flow in the network. 
We demonstrate that the proposed learning principle
allows targeted representations to be attained w.r.t. a local goal set only at one node located anywhere in the network.

\vspace{-0.1in}
\subsection{Notations and Paper Organization}
\vspace{-0.1in}
\textbf{Notations} A variable at node level $l$ has a subscript $*_l$. Scalars, vectors and matrices are denoted by usual, bold lower and bold upper case symbols as ${x}_l$, ${\bf x}_l$ and ${\bf X}_l$. A set of data samples from $C$ classes is denoted as ${\bf Y}_l=[{\bf Y}_{l,1},...,{\bf Y}_{l,C}] \in \Re^{M_l \times CK}$. Every class $c \in \{1,...,C\}$ has $K$ samples, ${\bf Y}_{l,c}$$=[{\bf y}_{l, \{c,1\} },...,{\bf y}_{l, \{c,K\}}] \in \Re^{M_l \times K}$. We denote the $k-$th representation from class $c$ at level $l$ as ${\bf y}_{l, \{c,k\}} \in \Re^{{M}_l}$, $\forall c \in \{1,...,C\}$, $\forall k \in \{1,...,K\}$, $\forall l \in \{1,...,L\}$.
The $\ell_p-$norm, nuclear norm, matrix trace and Hadamard product are denoted as $\Vert.\Vert_p$, $\Vert. \Vert_{*}$, $Tr()$ and $\odot$, respectively. 
The first order derivative of a function $\mathcal{L}({\bf Y}_{l})$ w.r.t. ${\bf Y}_l$ is denoted as $\frac{\partial \mathcal{L}({\bf Y}_{l})}{\partial {\bf Y}_l}$. We denote $\vert {\bf y}_{l, \{ c,k \}} \vert$ as the vector having as elements the absolute values of the corresponding elements in ${\bf y}_{l, \{ c,k \}}$. 

\textbf{Paper Organization} Section 2 introduces the generalized nonlinear transform, the local propagation modeling and explains the propagation dynamics. Section 3 gives connections to empirical risk, sets the learning target and presents the problem formulation. Section 4 proposes two learning algorithms that consider locally decoupled estimation per subsets of the network parameters and unveils new learning result. Section 5 devotes to numerical evaluation 
and Section 6 concludes the paper. 

\vspace{-0.1in}
\section{Network Description with Nonlinear  and Local Propagation Modeling}
\vspace{-0.1in}
\label{sec:ProblemFormulation}


In the first subsection, we introduce a generalized nonlinear transform modeling and 
in the second subsection we present and explain in detail 
the local propagation modeling. 



\vspace{-0.1in}

\subsection{Nonlinear Transform Model}
\vspace{-0.1in}

The 
data and the forward weights are denoted as ${\bf Y}_{0}=[{\bf y}_{0, \{1,1\}},...,{\bf y}_{0, \{C,K\}}] \in \Re^{N \times {CK} }$ and ${\bf A}_{l}\in \Re^{M_{l-1} \times M_{l}}, l \in \{1,...,L\}$, respectivly, where 
${\bf A}_{l-1}$ connects two nodes at levels $l-1$ and $l$.

\textbf{Sparsifying Nonlinear  Transform (sNT)} The sparse representation at node level $l$ defined w.r.t. 
a \textit{sparsifying transform} 
with parameter set $\mathcal{S}_l=\{ {\bf A}_{l-1}, \tau_{l} \}$ ($\tau_{l} \geq 0$ is a thresholding parameter) is denoted as:
\vspace{-.15in}
\begin{align}
&{\bf u}_{l,\{ c,k\}}={\rm sign}( {\bf q}_{l, \{c,k\}}) \odot \max( \vert {\bf q}_{l, \{c,k\}} \vert-\tau_{l}{\bf 1}, {\bf 0}), \label{eq:sparse:representation}
\end{align}
where ${\bf q}_{l, \{c,k\}}={\bf A}_{l-1}{\bf u}_{l-1, \{c,k\}}$ is the linear transform and ${\bf u}_{0, \{c,k\}}={\bf y}_{0, \{c,k\}}$.

\textbf{Generalized Nonlinear Transform (gNT)} Assume that 
positive thresholding vector ${\bf t}_{l, \{ c,k\}} \in \Re_{+}^{M_{l}}$, positive normalization vector ${\bf n}_{l, \{ c,k\}} \in \Re_{+}^{M_{l}}$, correction vector $\boldsymbol{\nu}_{l, \{ c,k\}} \in \Re^{M_{l}}$ and thresholding parameter $\lambda_{l, 1} \in \Re_+$ are given.
Denote ${\bf b}_{l, \{c,k\}}={\bf q}_{l, \{c,k\}}-\boldsymbol{\nu}_{l, \{ c,k\}}$ and ${\bf p}_{l, \{ c,k \}}={\bf t}_{l, \{ c,k\}}+\lambda_{l,1}{\bf 1}$, 
then the representation ${\bf y}_{l,\{c,k\}}$ at level $l$ defined w.r.t. the \textit{nonlinear transform} is:
\vspace{-.05in}
\begin{align}
&{\bf y}_{l,\{c,k\}}= {\rm sign}( {\bf b}_{l, \{c,k\}}) \odot \max( \vert {\bf b}_{l, \{c,k\}} \vert-{\bf p}_{l, \{ c,k \}}, {\bf 0})\oslash{\bf n}_{l, \{ c,k\}}. \label{eq:nonlinear:representation} 
\end{align}
The nature, the role and the interpretation of the variables ${\bf p}_{l, \{ c,k \}}, {\bf n}_{l, \{ c,k\}}$ and $\boldsymbol{\nu}_{l, \{ c,k\}}$ will be explained in details in the subsequent sections. For now, we refer to them as the portion of the total parameter set $\mathcal{P}_{l, \{c,k \}}={\{ {\bf A}_{l-1},  \{ {\bf p}_{l, \{ c,k \}}, {\bf n}_{l, \{ c,k\}}, \boldsymbol{\nu}_{l, \{ c,k\}} \} \}}$ that describes the nonlinear transform \eqref{eq:nonlinear:representation}. 
The 
transform \eqref{eq:nonlinear:representation} at node level $l$ is defined on top of the 
transform \eqref{eq:sparse:representation} at node level $l-1$.

\vspace{-.1in}

\vspace{-.06in}
\subsection{Local Propagation Model, Dynamics and Interpretations}
\label{LocalPropagatioModel:DynamicandInterpretations}
\vspace{-.1in}

This subsection first, defines the local goal for the representations at node level $l$, then introduces the local propagation term ${\mathcal{R}}_3(l)$,  gives its interpretation and explains its dynamics.

\textbf{Local Goal, Errors and Error Vectors} 
The local goal for the representations ${\bf U}_{l}=[{\bf u}_{l, \{1,1 \}},...,{\bf u}_{l, \{C,K \}}], l \in \{1,..,L\}$ at node level $l$ are the desired representations ${\bf G}_{l}=[{\bf g}_{l, \{1,1 \}},...,{\bf g}_{l, \{C,K \}}]$ that have specific properties. More formally, ${\bf G}_{l}$ are defined as the solution of an optimization problem where ${\bf G}_l$ has to be close to the linear transform representations ${\bf Q}_l={\bf A}_{l-1}{\bf U}_{l-1}$ while satisfying the constraints by $f_1$ and $f_2$,\textit{ i.e.}, $(P_{G}):\min_{{\bf G}_{l}}   \mathcal{L} ( {\bf Q}_l,{\bf G}_{l} )+f_1({\bf G}_l), \text{ subject to } f_2({\bf G}_{l})=0$, 
where $f_1,f_2: \Re^{M_{l} \times CK } \rightarrow \Re$ and $\mathcal{L}({\bf Q}_l, {\bf G}_l)=\frac{1}{2}\Vert {\bf Q}_{l}-{\bf G}_{l} \Vert_F^2$. 
Note that, in general, one might model different goals for the representations ${\bf U}_{l}$ w.r.t. the desirable properties by imposing constraints with a  properly defined functions $f_1$ and $f_2$. 

\vspace{-0.04in}
\textit{$-$ Discriminative and Sparse Representations}:  In this paper we use sparsity imposing constraint $f_1({\bf G}_l)=\mathcal{A}({\bf G}_l)=\lambda_{l,1}\sum_{c=1}^{C}\sum_{k=1}^{K} \Vert {\bf g}_{l,\{c,k\}} \Vert_1$ and 
knowing the corresponding labels 
a discrimination constraint $f_2({\bf G}_l)=\mathcal{U}({\bf G}_{l})=\lambda_{l,0}D({\bf G}_{l})=$  $\lambda_{l,0}\sum_{\icol{ c1, c1 \neq c} }\sum_{k1}( \Vert {\bf g}^{+}_{l,\{c,k\}}\odot{\bf g}^{+}_{l, \{ c1,k1\}} \Vert_1+$ $\Vert {\bf g}^{-}_{l, \{c,k \}}\odot{\bf g}^{-}_{l, \{c1,k1\}}\Vert_1$$+\Vert {\bf g}_{l,\{c,k\}}\odot{\bf g}_{l, \{ c1,k1\}} \Vert_2^2 )$, where ${\bf g}_{l, \{c,k\}}={\bf g}_{l, \{ c,k \}}^+-{\bf g}_{l, \{ c,k \}}^-$, ${{\bf g}_{l, \{c1,k1\} }^+=\max({\bf g}_{l, \{ c1, k1 \}}, {\bf 0})}$ and ${{\bf g}_{l, \{ c1,k1\}}^-=\max(-{\bf g}_{l ,\{ c1,k1\}}, {\bf 0})}$ 
\citet{Kostadinov:ICLR2018}. 
The solution of 
$(P_{G})$ for these particular functions 
is given in \textit{Appendix A.2}.

\vspace{-0.04in}

\textit{$-$ Local Goal Error, Nonlinear Transform Error and Error Vectors}: By considering the representations ${\bf G}_{l}$ and 
${\bf U}_{l}$ we define a \textit{goal error} (goal cost) as $\mathcal{G}({\bf G}_{l},{\bf U}_{l})=\frac{1}{2}\Vert {\bf G}_{l}-{\bf U}_{l} \Vert_F^2$. Similarly, by considering the representations ${\bf Q}_{l}$ and ${\bf Y}_{l}$ the \textit{nonlinear transform error} is $\mathcal{L}({\bf Q}_{l},{\bf Y}_{l})$. 
We distinguish two different error vectors at node level $l$. The first ones are associated to the local goal error (\textit{ge}) and the second ones are associated to the nonlinear transform error (\textit{te}). 
We define them as follows:
\vspace{-0.05in}
\begin{equation}
\begin{aligned}
\textit{ge:}&\textit{ } \frac{\partial \mathcal{G}({\bf G}_{l},{\bf U}_{l})}{\partial {\bf U}_{l}}=&& {\bf U}_{l}-{\bf G}_{l}, \textit{$ $  } \textit{$ $  } \textit{$ $  } \textit{$ $  } \textit{$ $  } \textit{$ $  } \textit{$ $  } \textit{$ $  } \textit{$ $  } \textit{$ $  }
\textit{te:}\textit{ } \frac{\partial \mathcal{L}({\bf Q}_{l},{\bf Y}_{l})}{\partial {\bf Y}_{l}}=&& {\bf Y}_{l}-{\bf Q}_{l}. \label{eq:error:vectors}
\end{aligned}
\end{equation}
The \textit{ge} vectors represent the deviation of the sparse representations ${\bf U}_{l}$ 
away from the ideal representations ${\bf G}_{l}$ and the \textit{te} vectors stands for the deviations in the nonlinear transform representations ${\bf Y}_l$ away from the linear transform representations ${\bf Q}_l={\bf A}_{l-1}{\bf U}_{l-1}$.


\vspace{-0.04in}
\textbf{Local Propagation Term} 
The term 
${\mathcal{R}}_3(l)$ 
is modeled as 
${\mathcal{R}}_3(l)=\lambda_{l,f}\mathcal{F}_{f}(l)+\lambda_{l,b}\mathcal{F}_{b}(l)$, 
where 
\vspace{-.0in}
{\small
 \begin{equation}
\begin{aligned}
\hspace{-0.21in}
\! \mathcal{F}_{b}(l)=\stackrel{}{Tr\left(\! \! \! \left( \! \frac{\partial \mathcal{L}({\bf Q}_{l},{\bf Y}_{l})}{\partial {\bf Y}_l} \right)^T \! \! \! \! \!  {\bf A}_{l-1} \! \frac{\partial \mathcal{G}({\bf G}_{l-1},{\bf U}_{l-1})}{\partial {\bf U}_{l-1}} \! \!  \right)}, \mathcal{F}_{f}(l)= \stackrel{}{Tr \!  \left( \! \! \!  \left(\! \! \frac{\partial \mathcal{L}({\bf Q}_{l},{\bf Y}_{l})}{\partial {\bf Y}_l}\right)^T \! \! \! \! {\bf B}_{l}\frac{\partial \mathcal{G}({\bf G}_{l+1},{\bf U}_{l+1})}{\partial {\bf U}_{l+1}} \right)},
\end{aligned}
\label{eq:global:problem:error:propagation:backward:and:forward}
\end{equation}}%
and $\lambda_{l,b}$ and $\lambda_{l,f}$ are regularization parameters. The first term in 
\eqref{eq:global:problem:error:propagation:backward:and:forward} regulates the inner product between the \textit{ge} vectors 
\eqref{eq:error:vectors} 
from the previous node, at level $l-1$, propagated through ${\bf A}_{l-1}$, and the \textit{te} vectors 
\eqref{eq:error:vectors} at the current node level $l$.
The second term in 
\eqref{eq:global:problem:error:propagation:backward:and:forward} regularizes the inner product between the \textit{ge} vectors 
\eqref{eq:error:vectors} 
at node level $l+1$, propagated through ${\bf B}_{l}$, and the \textit{te} vectors 
\eqref{eq:error:vectors} at node level $l$. 
\vspace{-0.04in}
\textbf{Local Propagation Dynamics and Interpretations} To explain the dynamics of the regularization, we start with the cases when both of the terms in 
\eqref{eq:global:problem:error:propagation:backward:and:forward} 
have no influence in the local 
model. The terms 
$\mathcal{F}_{f}$ and $\mathcal{F}_{b}$ 
will be zero if the \textit{ge} 
or the \textit{te} vectors 
\eqref{eq:error:vectors} are zero. In that case, either we achieve our local goal, since a sparse version of ${\bf Q}_{l-1}$ (or ${\bf A}_{l}{\bf U}_{l}$) 
equals the representations ${\bf G}_{l-1}$ (or ${\bf G}_{l+1}$), either ${\bf Q}_{l}$ equals\footnote{In general ${\bf A}_{l-1}{\bf U}_{l-1}$ is not sparse. However, it is possible 
${\bf A}_{l}{\bf U}_{l-1}$ to have any desirable properties within a very small error. 
} to the representations ${\bf Y}_{l}$ with the desired properties. 
The last case is when the affine combination between the propagated \textit{ge} 
vectors \eqref{eq:error:vectors} from node levels $l-1$ and $l+1$, through ${\bf A}_{l-1}$ and ${\bf B}_{l}$ 
are orthogonal to the \textit{te} vectors 
\eqref{eq:error:vectors}. To explain it, first, we give the connection to a local diffusion form. 
\vspace{-0.1in}
{\flushleft \textbf{Lemma 1} } \textit{By the fundamental lemma of the calculus of variations \citet{spivak1980calculus} and the conservation of energy \citet{Kittel:Charles_Kroemer:Herbert_1980}, 
$\mathcal{R}_{3}(l)$
has a diffusion 
 related form defined as}:
 \vspace{-0.066in}
\begin{equation}
\begin{aligned}
\hspace{-0.1in}\mathcal{R}_3(l) \backsimeq 
&Tr\left(\left(\frac{\partial \mathcal{L}({\bf Q}_{l},{\bf Y}_{l})}{\partial {\bf Y}_l}\right)^T  \nabla^2 \mathcal{G}({\bf U}_{l-1}, {\bf U}_{l+1} ) \right),
\end{aligned}
\label{eq:global:problem:error:propagation:flow}
\end{equation}
\vspace{-0.05in}
where ${\nabla^2 \mathcal{G}({\bf U}_{l-1}, {\bf U}_{l+1} ) }=\left[ \lambda_{l,f}{\bf B}_{l}\frac{\partial \mathcal{G}({\bf G}_{l+1},{\bf U}_{l+1})}{\partial {\bf U}_{l+1}}\right.+ \left.  \lambda_{l, b}{\bf A}_{l-1}\frac{\partial \mathcal{G}({\bf G}_{l-1},{\bf U}_{l-1})}{\partial {\bf U}_{l-1}} \right] $ is the local diffusion term, 
representing the vectors for the \textit{change of the local propagation flow}. 
They compactly describe 
the deviations of the representation ${\bf Y}_l$ w.r.t. the propagated \textit{ge} vectors 
$\frac{\partial \mathcal{G}({\bf G}_{l-1},{\bf U}_{l-1})}{\partial {\bf U}_{l-1}}$ and $\frac{\partial \mathcal{G}({\bf G}_{l+1},{\bf U}_{l+1})}{\partial {\bf U}_{l+1}}$
from node levels $l-1$ and $l+1$, 
through ${\bf A}_{l-1}$ and ${\bf B}_{l}$, respectively.

\vspace{-0.05in}
\textit{$-$ Preservation of the Change in the Goal Driven Local Propagation Flow}:  When 
${\nabla^2 \mathcal{G}({\bf U}_{l-1}, {\bf U}_{l+1} ) }$
is orthogonal to the transform error vectors $\frac{\partial \mathcal{L}({\bf Q}_l,{\bf Y}_{l})}{\partial {\bf Y}_l}$ at level $l$ it means that the \textit{change of the local propagation flow is preserved}. In other words, an alignment is achieved between the locally targeted 
nonlinear representations and the change of the goal driven propagation flow 
(Figure \ref{prior:dinimics:ilustration}). 

\vspace{-0.05in}
\textit{$-$ Reduction to Local Propagation Flow}  In the case that the local goal is zero, \textit{i.e.}, the representations ${\bf G}_{l-1}$ and ${\bf G}_{l+1}$ are zero vector, 
\eqref{eq:global:problem:error:propagation:flow} 
regularizes the \textit{local propagation flow} and 
takes the form as 
$\mathcal{R}_{3}(l)\backsimeq Tr\left((\frac{\partial \mathcal{L}({\bf Q}_l,{\bf Y}_{l})}{\partial {\bf Y}_l})^T  \nabla \mathcal{G}({\bf U}_{l-1}, {\bf U}_{l+1}) \right) $, where ${\nabla \mathcal{G}({\bf U}_{l-1}, {\bf U}_{l+1} ) }=\left[ \lambda_{l,f}{\bf B}_{l}{\bf U}_{l+1} \right.+ \left.  \lambda_{l, b}{\bf A}_{l-1}{\bf U}_{l-1} \right] $.     

\textit{$-$ Local Throughout and Entanglement}: Note that the nonlinear transform error at node level $l$ can be constrained in favor of the local propagation flow (or its change). 
The term \eqref{eq:global:problem:error:propagation:flow} reflects the ability of a network node at level $l$ to learn the properties of a desirable propagation flow (or its change) within network nodes at levels $l-1$ and $l+1$. 
We named it as \textit{local throughout}.
Taking into account both the local goal and the throughout gain at node level $l$, 
we have an implicit model for an \textit{entanglement} that influences on the representations at all nodes, globally in the network. 
The 
entanglement 
at one node in the network 
relates three entities, 
\begin{myitemize} 
\item[(i)] The model error $\mathcal{L}({\bf Q}_{l},{\bf Y}_{l})$ 
representing an information loss at node level $l$,
\item[(ii)] The strength of deviations $\Vert \nabla^2 \mathcal{G}({\bf U}_{l-1}, {\bf U}_{l+1} ) \Vert_F^2$ expressed through the change in the local propagation flow 
within
node levels $l-1$ and $l+1$, 
and
\item[(iii)]  The goal error 
$\mathcal{G}({\bf G}_{l},{\bf U}_{l})$ at node level $l$.
\end{myitemize}
The local throughout and the entanglement describes a trade-off where only the properties of the subset of the affected entities can be changed w.r.t. the properties of the rest by locally involving 
${\bf A}_{l-1},{\bf B}_{l}, {\bf Y}_l, {\bf Q}_{l}, {\bf U}_{l-1}, {\bf U}_{l}, {\bf U}_{l+1}$, ${\bf G}_{l-1}$ and ${\bf G}_{l+1}$\footnote{
Due to space limitations the information-theoretic \citet{Cover:2006:EIT:1146355} analysis with 
the precise and exact 
characterization of the fundamental limits in this trade-off are out of the scope of this paperer. 
}. An illustration is given on Figure \ref{prior:dinimics:ilustration}. 

\vspace{-0.15in}

\section{Network Learning with Local Propagation}
\label{sec:The:learning:algorithm}
\vspace{-0.1in}

This section explains the parametrization in the network operational modes, unveils the empirical risk as learning target and presents the general problem formulation for learning the network parameters with local propagation constraints. 

\vspace{-0.1in}
\subsection{Over-Parameterization, Empirical Risk and Learning Target}
\vspace{-0.1in}

We use two descriptions of all the network representations, one by gNT and the other by sNT. 

\vspace{-0.04in}
\textbf{gNT, sNT and Empirical Risk} 
The gNT and sNT per network node share the linear map ${\bf A}_l$, but, gNT have additional parameters. 

\vspace{-0.03in}
\textit{$-$ {Over-Parametrization with gNT}}: 
All representations ${\bf Y}_{l}$ at node level $l$ are modeled by a set of 
gNTs described by 
$\mathcal{P}_{l}=\{ \mathcal{P}_{l, \{ 1,1 \}}, ..., \mathcal{P}_{l \{C,K\}} \}$, 
where $\mathcal{P}_{l, \{c,k\}}=\{ {\bf A}_l, \{ {\bf p}_{l, \{ c,k \}}, {\bf n}_{l, \{ c,k\}}, \boldsymbol{\nu}_{l, \{ c,k\}} \} \}$. At node level $l$ the number of nonlinear transforms $\mathcal{P}_{l, \{c,k\}}$ equals to the number of the available training data samples that in our case is  $CK$, meaning that we have an over-parametrization with one gNT per one representation ${\bf y}_{l, \{c,k\} }$. All the nonlinear transforms $\mathcal{P}_{l, \{c,k\}}$ for node level $l$ share the linear map ${\bf A}_{l}$ and have different parameters $\{ {\bf p}_{l, \{ c,k \}}, {\bf n}_{l, \{ c,k\}}, \boldsymbol{\nu}_{l, \{ c,k\}} \}$. Different sets of nonliner transforms $\mathcal{P}_l$ are modeled across different node levels $l$. 
This is usefully since it allows to characterize the representations at any node under any constraints including the very important local propagation. As we will show in the following subsections, the local propagation component is explicitly identified and has an additively corrective role in the empirical risk.



\vspace{-0.03in}

\textit{$-$ {Simplification with sNT}}: 
All sNT representations ${\bf U}_{l}$ \eqref{eq:sparse:representation} at the node levels $l$ that are used during training and testing mode, in fact, represent a simplification to the over-parametrization by gNT.  

\vspace{-0.03in}
\textit{$-$ {Connecting gNT to sNT Through Empirical Risk}}: 
Let the sparsifying transform  ${\bf u}_{l, \{c,k\}}$ be given and $\tau_{l}=\lambda_{l,1}$, if $\boldsymbol{\nu}_{l,\{ c,k\}}=$${\bf t}_{l,\{ c,k\}}={\bf 0}$ and $ {\bf n}_{l, \{ c,k\}}={\bf 1}$ or if:
\begin{align}
\xi_{l, \{c,k \}}={\bf t}_{l,\{ c,k\}}^T\vert {\bf u}_{l,\{ c,k\}} \vert+{\boldsymbol{\nu}_{l,\{ c,k\}}^T}{\bf u}_{l,\{ c,k\}}+ {\bf n}_{l, \{ c,k\}}^T({\bf u}_{l,\{ c,k\}}\odot {\bf u}_{l,\{ c,k\}}), \label{eq:empirical:risk}
\end{align}
is zero, 
then 
\eqref{eq:nonlinear:representation} reduces to the sNT \eqref{eq:sparse:representation}. 
In general, 
$P_E:{\mathbb E}[ \xi_{l, \{c,k \}} ] \simeq \frac{1}{CK}\sum_{c,k} \xi_{l, \{c,k \}}$,
can be seen as an empirical risk for the sNT representations ${\bf U}_{l}$ 
and the corresponding sparsifying model with parameter set $\mathcal{S}_l$. 
Meaning that any ${\bf y}_{l, \{ c,k\}}$ from ${\bf Y}_{l}=[{\bf y}_{l, \{1,1\}}, ..., {\bf y}_{l, \{C,K\}}] \in \Re^{M_{l} \times {CK} }$ 
can be analyzed using the corresponding ${\bf u}_{l, \{c,k\}}$  and its empirical risk $\xi_{l, \{c,k \}}$. 

\vspace{-0.03in}
\textbf{Learning Target} 
In the learning mode we target to estimate the parameter set $\{ \mathcal{S}_{1}, ..., \mathcal{S}_{L}\}$ for the sNTs that approximate the parameter set $\{ \mathcal{P}_{1},...,\mathcal{P}_{L} \}$ of the gNTs. One sNT defined by $\mathcal{S}_l=\{ {\bf A}_l, \tau_{l} \}$ approximates one set of gNTs defined by $\mathcal{P}_l=\{ \mathcal{P}_{l, \{1,1\}},..., \mathcal{P}_{l, \{C,K\}}\} \}$.  In other words, for every node at level $l$, given ${\tau_l}$, we would like to estimate ${\bf A}_l$ for the sNT \eqref{eq:sparse:representation} that minimize the empirical risk $(P_E)$ 
. By doing so, after the learning mode is finished, we would like 
${\bf Y}_{l}$ to be 
equal to 
${\bf U}_{l}$ that is used at testing time. 

\vspace{-0.1in}
\subsection{Problem Formulation}
\vspace{-0.1in}

The learning of the network parameters 
is addressed by the following problem formulation:
\vspace{-0.12in}
\begin{align}
\boldsymbol{\Omega}&={\rm arg} \min_{\boldsymbol{\Omega} } 
\sum_{l=1}^L \left( \mathcal{R}_{1}(l)+\mathcal{R}_{2}(l)+\mathcal{R}_{3}(l)+\mathcal{A}({\bf Y}_{l}) \right) +\mathcal{U}({\bf Y}_{l3}), \label{eq:global:problem:formulation} \text{where}
\end{align}
\vspace{-0.25in}
\begin{equation}
\begin{aligned}
\hspace{-0.0in} 
\mathcal{R}_{1}(l)&= \mathcal{L}({\bf A}_{l-1}{\bf U}_{l-1}, {\bf Y}_{l})+\mathcal{L}({\bf B}_{l}{\bf U}_{l+1}, {\bf Y}_{l}), \text{${ }$ } \text{${ }$ } \text{${ }$ } \text{${ }$ } 
\mathcal{R}_{2}(l)=\mathcal{V}({\bf A}_{l-1})+
\sum_{l2=l-1}^{l}\mathcal{W}({\bf A}_{l2}, {\bf B}_{l2}), \label{eq:global:problem:formulation:terms:a} \nonumber 
\end{aligned}
\end{equation}
and  $\boldsymbol{\Omega}=\{ {\bf A}_0,., {\bf A}_{L-1}, {\bf U}_1,.,{\bf U}_{L}, {{\bf Y}_1,.,{\bf Y}_{L}, {\bf B}_0,., {\bf B}_{L-1}} \}$ are the network parameters and $l_3 \in \{1,...,L\}$. The term $\mathcal{R}_1(l)$ models the representations at node level $l$, the term $\mathcal{R}_2(l)$ models the properties of the weights that connect nodes at levels $l-1$ and $l$, as well as 
nodes at levels $l$ and $l+1$, and the term $\mathcal{R}_{3}(l)$ models the local propagation at node level $l$ 
from the nodes at levels $l-1$ 
and $l+1$.
%
The 
backward weighs are denoted as ${\bf B}_{0}\in \Re^{N \times M_1}, {\bf B}_{l}\in \Re^{M_{l} \times M_{l+1}}$.  We introduce them to be able to reconstruct ${\bf Y}_{l}$ at layer $l$ from  ${\bf Y}_{l+1}$ at layer $l+1$, to avoid solving a constrained inverse problem related to ${\bf Y}_l$ and regularize the local propagation. Instead of ${\bf B}_l$, we can also use ${\bf A}^T_l$. To present the full potential of the approach, we consider a general case with ${\bf B}_l$ different from ${\bf A}_l^T$.

At all nodes, the representations are constrained to be sparse by $\mathcal{A}({\bf Y}_l)=\lambda_{l,1}\sum_{c=1}^{C}\sum_{k=1}^{K} \Vert {\bf y}_{l,\{c,k\}} \Vert_1$. 
At each node, the corresponding representations ${\bf Y}_{l}$ can be constrained by $\mathcal{U}({\bf Y}_{l})$ to have specific properties. This paper focuses on the discriminative properties of the representations for a particular node at level $l3 \in \{1,..,L\}$. 
The terms $\mathcal{V}({\bf A}_{l-1})=\frac{\lambda_{l,2}}{2}\Vert {\bf A}_{l-1} \Vert_{F}^2+\frac{\lambda_{l,3}}{2}\Vert{\bf A}_{l-1}{\bf A}_{l-1}^T-{\bf I} \Vert_F^2-\lambda_{l,4}\log \vert \det {\bf A}_{l-1}^T{\bf A}_{l-1} \vert$ and $\mathcal{W}({\bf A}_{l-1}, {\bf B}_{l-1})=\frac{\lambda_{l,5}}{2}\Vert {\bf A}_{l-1}-{\bf B}_{l-1}^T \Vert_F^2$ 
are used to regularize the conditioning, the expected coherence 
of ${\bf A}_{l}$ \citet{SPARS:Kostadinov2017}, and the similarity between ${\bf A}_{l}$ and ${\bf B}_{l}^T$.  
\vspace{-0.1in}
\subsection{Synchronous and Asynchronous Learning Algorithms with Parallel Execution}
\vspace{-0.1in}




This section presents 
the solution to \eqref{eq:global:problem:formulation} 
using two learning strategies, each with two stages.
\textbf{Learning Strategy} Two stages are responsible for estimating (i) the resulting sNT representations and (ii) the resulting gNT representations and the corresponding linear maps ${\bf A}_{l-1}$ and ${\bf B}_l$.  Concerning the estimation of ${\bf G}_l$, we have three choices:
\begin{minipage}[t]{1\textwidth}
\vskip -0.05in
\centering

(i) no goal, it corresponds to ${\bf G}_l={\bf 0}$, \text{${ }$ } \text{${ }$ } \text{${ }$ } \text{${ }$ } \text{${ }$ } \text{${ }$ }  \text{${ }$ }  \text{${ }$ } \text{${ }$ }  \text{${ }$ }  \text{${ }$ }  \text{${ }$ }  \text{${ }$ }  \text{${ }$ }(ii) predefined and fixed goal and

\text{${ }$ } (iii) dynamic goal by computing ${\bf G}_l$ w.r.t. ${\bf A}_{l-1}{\bf U}_{l-1}$, $f_1$ and $f_2$.
\end{minipage}

\textit{$-$ Stage One}: This stage computes a forward (or backward) pass through the network using the weighs ${\bf A}_{l}$ (or the weighs ${\bf B}_l$). 
We explain two possible setups for execution of this stage.
In the first setup, a hold is active till all weights ${\bf A}_{l}$ (or ${\bf B}_{l}$) in the network are updated. Afterwards, the execution proceeds, which corresponds to a \textit{synchronous} case. 
In the second setup, at one point in time, one takes all the available weights ${\bf A}_{l}$ (or ${\bf B}_{l}$), whether updated or not, and then executes the first stage, which corresponds to an \textit{asynchronous} case. 
In this stage the representation ${\bf U}_{l}$ at one node is considered as a sNT \eqref{eq:sparse:representation} on the representation from the previous (or next) node. 

\vspace{-0.07in}
\textit{$-$ Stage Two}: (\textit{Locally Decoupled Parallel Mode})
Second stage computes a parallel update on all subsets $\varsigma_l=\{ {\bf Y}_l, {\bf A}_{l-1}, {\bf B}_{l} \}$ of network parameters. 
Every subset $\varsigma_l$ forms a local network consisting of a node and its directly connected neighboring nodes with propagation directed to that node. In this stage, the representations ${\bf Y}_l$ at one node are considered as a gNT \eqref{eq:nonlinear:representation} applied to the sNT representations from the previous node. The local network is shown in Figure \ref{intro:image:prior:dynamics}. 
Common learning problem with regularized local propagation addresses the parallel updates on all $\varsigma_l$.
All the subproblems for the corresponding $\varsigma_l$ are decoupled. 
In the corresponding subproblems any $\varsigma_{l1}$ does not share parameters with any other $\varsigma_{l2}$, \textit{i.e.}, $\varsigma_{l1} \bigcap \varsigma_{l2}=\emptyset, \forall l1 \neq l2$. 

\vspace{-0.07in}
The desired representations ${\bf G}_l$ can be computed in stage one or two. At stage one the corresponding ${\bf A}_{l-1}$ is given, therefore, after estimating ${\bf U}_{l-1}$, ${\bf G}_l$ can be estimated. At stage two ${\bf U}_{l-1}$ is given, therefore, after estimating ${\bf A}_{l-1}$, ${\bf G}_l$ can be estimated.
\vspace{-0.03in}
\textbf{Locally Decoupled Learning} 
Let all the variables in problem \eqref{eq:global:problem:formulation} be fixed except $\varsigma_l$ 
then \eqref{eq:global:problem:formulation} 
reduces to the following problem:  
\vspace{-0.1in}
\begin{equation}
\begin{aligned}
\min_{ \varsigma_l}{\mathcal{R}}_{1}(l)+{\mathcal{R}}_{2}(l)+{\mathcal{R}}_{3}(l)+{\mathcal{A}}({\bf Y}_l)+{\mathcal{U}}({\bf Y}_l),
\end{aligned}
\label{eq:global:problem:formulation:reduction}
\end{equation}
where the network allows the possibility per node level $l$, to define different ${\mathcal{A}}({\bf Y}_l)$ and ${\mathcal{U}}({\bf Y}_l)$, or not consider them. 
Problem \eqref{eq:global:problem:formulation:reduction} is still non-convex. Nevertheless, to solve \eqref{eq:global:problem:formulation:reduction}, we propose an alternating block coordinate descend algorithm where we iteratively update one variable from the set of variables $\varsigma_l=\{ {\bf Y}_l, {\bf A}_{l-1}, {\bf B}_{l} \}$ while keeping the rest fixed. 
It has three steps: (i) 
estimation of the representation ${\bf Y}_l$ with local propagation, 
(ii) and (iii) estimation of the forward and backward weights ${\bf A}_{l-1}$ and ${\bf B}_{l}$, respectively, with local propagation. In the following we explain the steps of the proposed solution, identify the empirical risk and highlight the involved trade-offs. 

\textit{\underline{$-$ Representation Estimation with Local Propagation}} 
\label{label:sec:representation:estimate}
Let all the variables in problem \eqref{eq:global:problem:formulation:reduction} be given except ${\bf y}={\bf y}_{l, \{c,k\}}$ then it 
reduces to the following \textit{constrained projection problem}:
\vspace{-0.1in}
\begin{align}
&\min_{ \icol{{\bf y} } }\frac{1}{2}\Vert {\bf x}-{\bf y} \Vert_2^2+{\boldsymbol{\nu}^T}{\bf y}+(\lambda_{l,1}{\bf 1}+{\bf t})^T\vert {\bf y} \vert+{\bf n}^T({\bf y}\odot {\bf y}), \label{eq:global:problem:formulation:reduction:representation:estimation} 
\end{align}
we dropped the subscripts $_{l \{c,k \}}$ to simplify notation, ${\bf x}={\bf x}_{l, \{c,k\}}=
{\bf A}_{l-1}{\bf u}_{l-1, \{c,k\}}+{\bf B}_{l}{\bf u}_{l+1, \{c,k\}},  {\bf t}=\lambda_{l, 0}( {\rm sign}({\bf w}^+)\odot{\bf d}^+ +{\rm sign}({\bf w}^-)\odot{\bf d}^-)$, $\boldsymbol{\nu}=\boldsymbol{\nu}_{l, \{c,k\}}=$ $\lambda_{l,f} {\bf B}_{l} \left({\bf u}_{l+1, \{c,k\}}-{\bf g}_{l+1, \{c,k\}} \right)-\lambda_{l,b} {\bf A}_{l-1}\left({\bf u}_{l-1, \{c,k\}}-{\bf g}_{l-1, \{c,k\}} \right),$
${\bf d}^- =\sum_{\icol{ c1, c1 \neq c} }\sum_{k1}{\bf w}_{l, \{c1,k1 \}}^{-}, {\bf d}^+ =\sum_{{ c1,\ c1 \neq c} }\sum_{k1}{\bf w}_{l, \{c1,k1 \}}^{+}$, ${\bf n}=\sum_{\icol{ c1, c1 \neq c} }\sum_{k1}{\bf w}_{l, \{c1,k1 \}} \odot {\bf w}_{l, \{c1,k1 \}}$ and $ {\bf w}={\bf w}_{l, \{c1,k1 \}}={\bf x}_{l, \{c,k\}}-\boldsymbol{\nu}_{l, \{c,k\}}$. Problem \eqref{eq:global:problem:formulation:reduction:representation:estimation} 
has a closed form solution as: 
\vspace{-0.05in}
\begin{equation}
\begin{aligned}
{\bf y}={\rm sign}({\bf x}-\boldsymbol{\nu} ) \odot\left(  \vert {\bf x}-\boldsymbol{\nu} \vert -{\bf t}-\lambda_{l,1}{\bf 1}, {\bf 0} \right)\oslash{\bf n}.
\end{aligned}
\label{eq:global:problem:formulation:reduction:representation:estimation:solution}
\end{equation}
The proof is given in \textit{Appendix A}. By \eqref{eq:global:problem:formulation:reduction:representation:estimation:solution} all ${\bf y}_{l, \{c,k\}}$ at node level $l$ can be computed in parallel. 

\vspace{-.1in}
\textit{ } ${ }$ \fbox{\begin{minipage}{6em}
{\textit{Empirical Risk}}
\end{minipage}} Denote ${\bf u}={\bf u}_{l, \{c,k\}}$, if ${\bf t}$ and $\boldsymbol{\nu}$ are zero and ${\bf n}$ is one or if
${\xi}_{l, \{c,K\}}$ \eqref{eq:empirical:risk}
is zero then \eqref{eq:global:problem:formulation:reduction:representation:estimation:solution} reduces to the sNT \eqref{eq:sparse:representation}. In addition, note that the empirical expectation 
${\mathbb E}[{\xi}_{l, \{c,K\}}]$ 
exactly matches the constraints ${\mathcal{U}}({\bf Y}_{l})+\mathcal{R}_{3}(l)$, \textit{i.e.}, 
${\mathbb E}[{\xi}_{l, \{c,K\}}] $
$\sim \frac{1}{CK}\left( {\mathcal{U}}({\bf Y}_{l})+\mathcal{R}_{3}(l)\right)$ 
and indicates them as {empirical risk} for the sNT \eqref{eq:sparse:representation}. 

Note that if 
${\mathcal{U}}({\bf Y}_l)$ is not used, 
${\bf t}={\bf n}={\bf 0}$, if 
${\mathcal{A}}({\bf Y}_l)$ is not used, 
${\lambda_{l,1}}={ 0}$ and by not using both 
\eqref{eq:global:problem:formulation:reduction:representation:estimation:solution} simplifies as ${\bf y}={\bf x}-\boldsymbol{\nu}$. In addition if we reorder ${\bf x}-\boldsymbol{\nu}$ 
then we have ${\bf x}-\boldsymbol{\nu}={\bf B}_{l}( (1+\lambda_{l,f}) {\bf u}_{l+1}-\lambda_{l,f}{\bf g}_{l+1})-{\bf A}_{l-1}( (\lambda_{l,b}- 1){\bf u}_{l-1}-\lambda_{l,b}{\bf g}_{l-1})$, 
which give us a 
precise view about the influence of the local propagation form node levels $l-1$ and $l+1$ in the estimation of the representation ${\bf y}_{l, \{ c,k \} }$ at node level $l$. 
It is completely 
determent by the sparse representations ${\bf u}_{l-1}$ and ${\bf u}_{l+1}$, their goal related representations ${\bf g}_{l-1}$ and ${\bf g}_{l+1}$, the parameters $\lambda_{l,f}, \lambda_{l,b}$ and the weighs ${\bf A}_{l-1}$ and ${\bf B}_{l}$. 

\vspace{-.1in}
\textit{ } ${ }$ \fbox{\begin{minipage}{4.2em}
{\textit{Trade-Off}}
\end{minipage}} Problem \eqref{eq:global:problem:formulation:reduction:representation:estimation} 
captures a trade-off between:
a) representations discrimination b) influence by the local propagation constraint and
c) reconstruction from level $l+1$ to level $l$.
\textit{\underline{$-$ Forward Weights Estimation with Local Propagation}} 
Let all the variables in problem \eqref{eq:global:problem:formulation:reduction} be given except ${\bf A}_{l-1}$ 
then \eqref{eq:global:problem:formulation:reduction} 
reduces to the following problem $(P_{FWE}):\min_{ \icol{{\bf A}_{l} } } \mathcal{R}_1(l)+\mathcal{R}_2(l)+{\mathcal{R}}_3(l)$. 
Denote ${\bf R}_L=\lambda_{l,f}{\bf Z}_{f,e}^T{\bf U}_{l-1}+ \frac{\lambda_{l,5}}{2}{\bf B}_{l}  +\gamma {\bf U}_{l-1}{\bf V}_{l}^T$ and ${\bf R}_Q=\frac{\pi}{2}{\bf I}+
 (\frac{\lambda_{l,b}}{2} {\bf Z}_{b,e}+\frac{\gamma^2}{2}{\bf U}_{l-1})({\bf U}_{l-1})^T$
where the terms ${\bf Z}_{b,e}=\frac{\partial \mathcal{G}({\bf G}_{l-1},{\bf U}_{l-1})}{\partial {\bf U}_{l-1}}$, ${\bf Z}_{f,e}={\bf B}_{l}\frac{\partial \mathcal{G}({\bf G}_{l+1},{\bf U}_{l+1})}{\partial {\bf U}_{l+1}}$, $\pi=\lambda_{l,2}+\lambda_{l,5}-\lambda_{l,3}$ and $\gamma=\kappa\eta$.
Assuming that the eigen value decomposition ${\bf U}_X{\bf \Sigma}_X^2{\bf U}_X^T$ of ${\bf R}_Q$ and the singular value decomposition ${\bf U}_{U_XXY}{\bf \Sigma}_{U_XXY}{\bf V}^T_{U_XXY} $ of ${\bf R}_L$ exist
then if and only if ${\Sigma}_{X}(n,n)={\sigma}_{X}(n) \geq 0$, $\forall n \in \{ 1,...,M_{l-1}\}$, 
$(P_{FWE})$ 
has approximate closed form solution as: ${\bf A}_{l-1}={\bf V}_{U_XXY}{\bf U}_{U_XXY}^T{{\bf \Sigma}_A}{{\bf \Sigma}_X^{-1}}{\bf U}_X^T$, 
{where ${\bf \Sigma}_A$ is diagonal matrix, ${\Sigma}_A(n,n)$ $={\sigma}_A(n) \geq 0$, and $\sigma_A(n)$ are solutions to a quartic euqation} 
(\textit{the proof is given in Appendix D}). 
Note that if ${\bf A}_{l-1}$ is under-complete, square or square orthogonal 
the related problem 
has a closed form solution. 


\vspace{-.1in}
\textit{ } ${ }$ \fbox{\begin{minipage}{4.2em}
{\textit{Trade-Off}}
\end{minipage}} In this step the trade-off is between:
%
a) the similarity of ${\bf A}_{l-1}$ to ${\bf B}_{l-1}^T$,
b) the influence of the local propagation and 
c) the influence of the local goal in the estimate of ${\bf A}_{l-1}$.
\textit{\underline{$-$ Backward Weight Estimation with Local Propagation}}
Let all the variables in problem \eqref{eq:global:problem:formulation:reduction} be given except ${\bf B}_{l}$ 
then \eqref{eq:global:problem:formulation:reduction} reduces to the following problem $(P_{BWE}):\min_{ \icol{{\bf B}_{l} } } \mathcal{R}_1(l)+\mathcal{R}_2(l)+{\mathcal{R}}_3(l)$.  
Denote ${\bf L}={\bf Q}_{l}{\bf U}_{l+1}^T+\lambda_{l,5}{\bf A}_{l}^T$ $-\lambda_{l-1,f}( {\bf G}_{l+1}-$ ${\bf U}_{l+1} ){\bf Y}_{l}^T$, then 
$(P_{BWE})$
has a closed form solution as: 
${\bf B}_{l}={\bf L}\left({\bf U}_{l+1}({\bf U}_{l+1})^T+\lambda_{l,5}{\bf I}\right)^{-1}.$
Note that if the weights ${\bf A}_{l}$ are orthonormal, then there is no need for an additional backward weights ${\bf B}_{l}$, since ${\bf B}_{l}={\bf A}_{l}^T$. 

\vspace{-.1in}
\textit{ } ${ }$ \fbox{\begin{minipage}{4.2em}
{\textit{Trade-Off}}
\end{minipage}} Moreover, the trade-off is between: a) how close is ${\bf B}_l$ to ${\bf A}_l^T$, b) how strong is the local back propagation and c) the accuracy of reconstruction $\Vert {\bf B}_{l}{\bf U}_{l+1}-{\bf Y}_{l} \Vert_F^2$.

%
%


\vspace{-0.12in}
\subsection{Nontrivial Local Minimum Solution Guarantee}
\vspace{-0.1in}
The next result shows that with arbitrarily small error 
we can find a 
local minimum solution to 
\eqref{eq:global:problem:formulation}. 
\textbf{Theorem 1}
\textit{Given any data set ${\bf Y}_0$, there exists} $\boldsymbol{\omega} =\{ \lambda_{1,bf}, ... , \lambda_{L,bf} \}, \lambda_{l, bf}=\{ \lambda_{l,b}, {\lambda_{l,f}}{} \}$
\textit{$\lambda_{l,b} >0, \lambda_{l,f}>0$ and a learning algorithm for a $L$-node transform-based network with a 
goal set on one 
node at level $l_G$ 
such that the 
algorithm 
after $t > S$ iteration 
learns all ${\bf A}_{l}, {l} \in \{0,...,L-1\}$ 
with}
\textit{
$\mathcal{G}({\bf D}_{L},{\bf U}_{L})  = {\epsilon}$, where  ${\bf D}_{L} \in \Re^{M_L \times CK}$ are the resulting representations 
of the propagated ideal representations ${\bf G}_{l_G}$ through the network 
from node level $l_G+1$, and ${\epsilon}>0$ is arbitrarily small constant}. 
The proof is given in \textit{Appendix E}.

\textbf{Remark} 
\textbf{} \textit{ The result by \textbf{Theorem 1} reveals the possibility to attain 
desirable 
representations ${\bf U}_{L}$ at level $L$ 
while only setting one local representation goal on one node at level $l_G \in \{1,..,L \}$. } 
\textit{Any network equipped with the proposed nonlinear and local propagation modeling, a goal expressed with $f_1$ and $f_2$, and properly chosen ${\boldsymbol{\omega}}$ can be trained using the proposed algorithm.}

\vspace{-0.05in}

\vspace{-0.15in}
\section{Evaluation of Recognition Accuracy}

\vspace{-0.15in}


This section evaluates the proposed local learning strategy.

\vspace{-0.15in}
\subsection{Data Set and Evaluated Networks} 
\vspace{-0.1in}
\textbf{Data Sets}  We present preliminary results using the a feed forward network with ${\bf B}_l={\bf A}_l^T$ in a supervised setup. The used data sets are MNIST 
and Fashon-MNIST. 
All the images from the data sets are downscaled to resolution 
$28\times28$, and are normalized to unit variance. 

\vspace{-0.07in}
\textbf{Networks}  A summary of the considered supervised cases 
is as follows. For the MNIST and the F-MNIST database we analyse $12$ different networks, $6$ per database. Per one database $4$ networks have $6$ nodes and aditional $2$ have $4$ nodes. The networks are trained in synchronous \textit{syn} and asynchronous mode \textit{asyn}. For the $6$-node networks trained in \textit{syn}, $2$ of them have a goal defined at the last node $L$ (\textit{syn}-$n[6]g[6]$) and for the remianing $2$ the goal is set on node at the middle in the network at level $3$ (\textit{syn}-$n[6]g[3]$). For the $4$-node network the goal is set at node level $4$ (\textit{syn}-$n[4]g[4]$). Simmilary in \textit{asyn} mode we denote the networks as (\textit{asyn}-$n[6]g[6]$), (\textit{asyn}-$n[6]g[3]$) and (\textit{asyn}-$n[4]g[4]$).      

\vspace{-0.15in}
\subsection{Learning and Testing Setup} 
\vspace{-0.07in}
\textbf{Learning setup} 
The training is done as explained in section 3.1. 
The asynchronous mode is implemented by using $L$ random draws $\boldsymbol{\psi} \in \{-1,1\}^{L}$, as the number of nodes, from a Bernoulli distribution. 
If the random 
realization is $1$, $\psi(l)=1$, we use ${\bf A}_l^{t}$ in the forward pass (stage one) and we update the corresponding set of varibles $\varsigma_l$ (stage two). If the random 
realization is $-1$, $\psi(l)=-1$, then we do not use ${\bf A}_l^{t}$
, but, instead we use ${\bf A}_l^{t-1}$ for stage one and in stage two we do not update the corresponding set $\varsigma_l$. 
A batch,  \textit{on-line} variant for weights update is defined as ${\bf A}^{t+1}_l={\bf A}_l^{t}-\rho({\bf A}^{t}_l-{{\bf A}}_l^{t-1} )$ 
where 
${{\bf A}}_l^t$ and ${{\bf A}}_l^{t-1}$ are the solutions to $(P_{FWE})$ w.r.t. subsets of the available training set at time steps $t$ and $t-1$, 
and $\rho$ is a predefined step size (
the details are given in \textit{Appendix D.3}). 
The batch portion equals to $15\%$ of the total amount of the available training data. 
The parameters $\{\lambda_{l,1}, \lambda_{l,2}, \lambda_{l,3}, \lambda_{l,4}, \lambda_{l,5} \} = \{34, 34, 34, 34, 34 \}$ and $\lambda_{l,1}=M_l/(2 \times l)$.
All the parameters $\lambda_{l,fb}$ are set as $\lambda_{l,fb}=\{1, 1\}$. 
The algorithm is initialized with a random matrices having i.i.d. Gaussian (zero mean, variance one) entries and is terminated after $120$ iterations. 

\vspace{-0.07in}
\textbf{Testing setup} 
During recognition 
the training data are propagated through the network using the sparsifying transform till node $L-1$, on node $L$ the nonlinear transform is computed, where $L$ is the last node. The test data are propagated through the network using the sparsifying transform till node $L$. 
The recognition results are obtained 
by using the test network representations and a $k-$NN search \citet{Cover:2006:EIT:1146355} over the training nonlinear transform representations. 
Using the MNIST database, the results are as follows \textit{syn}-$n[4]g[4]$, \textit{syn}-$n[6]g[6]$, \textit{syn}-$n[6]g[3]$ achieve accuracy of $\{ 98.1, 99.3, 97.6 \}$ and  \textit{asyn}-$n[4]g[4]$, \textit{asyn}-$n[6]g[6]$, \textit{asyn}-$n[6]g[3]$ achieve accuracy of $\{ 97.3, 98.7, 96.8. \}$, respectively. Using the F-MNIST database, the results are as follows \textit{syn}-$n[4]g[4]$, \textit{syn}-$n[6]g[6]$, \textit{syn}-$n[6]g[3]$ achive acuracy of $\{ 92.1, 93.1, 91.6 \}$ and  \textit{asyn}-$n[4]g[4]$, \textit{asyn}-$n[6]g[6]$, \textit{asyn}-$n[6]g[3]$ achieve accuracy of $\{ 91.3, 92.1, 90.8. \}$, respectively.

\vspace{-0.07in}
\textbf{Evaluation Summary} At this stage, the networks trained using the proposed algorithms on both of the used databases achieve comparable to state-of-the-art recognition performance \citep{pmlr-v28-wan13}. The networks have very small number of parameters, \textit{i.e.}, $6$ networks with $6$ nodes having $6$ weights with dimensionality  $784 \times 784$ and $4$ networks with $4$ nodes having $4$ weights with dimensionality $784 \times 784$. The learning time for $L=6$ node network is $12$ hours (on a PC that has Intel® Xeon(R) 3.60GHz CPU and 32G RAM memory) using a not optimized Matlab code that implements the sequential variant of the proposed algorithm. We expect a parallel implementation of the proposed algorithm to provide $\sim L \times$ speedup, which would reduce the learning time to $\sim 2$ hours.

\vspace{-0.15in}
\section{Discussion}
\vspace{-0.15in}
This paper was centered on novel joint treatment of three  modeling concepts: nonlinear transform, local goal and local propagation. 
It offered new results with intuitive interpretation and unfolded unified understanding of the learning dynamics for any network. 
Theoretically and by a numerical simulation, we showed that by learning locally (per node) with a local propagation constraint we can achieve targeted representations at the last node in the network. 
The preliminary evaluation 
was promising. On the used 
databases, the feed-forward network trained using the proposed algorithm with supervision 
provided comparable to state-of-the-art 
results. 

In the following we explain relations and show how the existing networks and their learning principles can be seen as special cases of the proposed principle with constrained local propagation.

\subsection{Connections to Existing Deep Neural Networks}  
The learning principle considering the regularization of the change in the propagation flow represents a generalization of the most known learning principles for learning in multilayer neural networks, but, in a localized manner. That is w.r.t. the parameters in a local network consisting of a node, its directly connected neighboring nodes through weights and the weights themselves. 


We show that the learning principles behind Local Forward/Back Propagation (FP/BP), Residual Network (RN)\footnote{Two cases are covered backward and forward difference. That is based on the difference of the representation at one node and the representation at a lower or upper node level.} \citep{HeZRS15}, Auto Encoder (AE) \cite{Baldi:2011AE} and its denoising extension \cite{VincentLLBM10},  Fully Visible Boltzmann Machines (FV-BM), Restricted Boltzmann Machines (RBM) \citep{Hinton:2017BoltzmannMachines}, Convolutional Neural Networks (CNNs) \citet{LeCunBH:2015:DL}, Recurrent Neural Networks (R-NN) \cite{Schmidhuber:2014:Overview}, \citet{Lipton:2015RNN}, Adversarial Networks (AN) \cite{Goodfellow:2014AN}, \cite{Makhzani:2015AAE}, \cite{YunchenPu:2017SVA_AN}  and Variational Auto-encoder (VAE) \citet{KingmaW2013:VAE, arjovsky:2017WAE} are just a special cases and a particular reductions of the proposed principle with regularization on the change in the local propagation flow. That is all of them not-explicitly constrain the propagation flow during the learning and try to achieve a propagation flow with a certain properties, but, under different goals and therefore, using different regularizes. 

\textbf{Local FP/BP}
Let $\lambda_{l,f}=0$ (or $\lambda_{l,b}=0$) be a particular choice of the parameters $\lambda_{l, b}$ and $\kappa$ (or $\lambda_{l, f}$ and $\kappa$) then the estimate of the representation at node level $n$ is regularized by a local back-propagation from node level $l+1$ (or by a local forward-propagation from node level $l-1$), that in turn is the gradient (the goal error) of the local goal at node level $l+1$ (or node level $l-1$).

\textbf{RN}
On the other hand note that if $\lambda_{l,f}=0$, then by using specific values for the parameters $\lambda_{l, b}$ and $\kappa$ the representation at node level $l$ actually encodes the differences between two transform representations at levels $l-1$ and $l$ and might be viewed as a residual network. It is the same principle if $\lambda_{l,b}=0$, then under other choice of the parameters $\lambda_{l, f}$, the proposed network again reduces to a residual network for two transform representations at node levels $l$ and $l+1$.

\textbf{AE}
In the proposed network, the forward pass ${\bf A}_{l-1}{\bf U}_{l-1}$ is the local encoder and the backward pass ${\bf B}_{l}{\bf U}_{l+1}$ is the local decoder, leading to an encoder-decoder pair. Note that at the same time, any goal can be set for the network. Additionally, if the goal is related to the reconstruction, then different trade-offs come to light. That is if ${\bf G}_{l-1}={\bf B}_{l-1}{\bf G}_{l}$, again ${\bf A}_{l-1}$ represents a forward projection operator and ${\bf B}_{l-1}$ represents a backward reconstruction operator and the connection to the auto-encoder is evident, for $\lambda_{l,f}=0$ (or $\lambda_{l,b}=0$). Furthermore, if the transform model is replaced with a synthesis model \citep{elad07:analysisvsynthesis} then the regularization of the propagation flow is w.r.t. the \textit{reconstruction error}. 

\textbf{RBM}
Although RBM models a probability distribution we give the connection w.r.t. its free energy. If the goal is zero as it was described in section \ref{LocalPropagatioModel:DynamicandInterpretations} the regularization term breaks down to the propagation flow, since $\nabla^2 \mathcal{G}({\bf G}_{l-1}, {\bf G}_{l+1})=\nabla \mathcal{G}({\bf U}_{l-1}, {\bf U}_{l+1})$. Moreover, in that case the proposed regularization term can be seen as a product of two experts \cite{Hinton:2002:TPE} one a fully visible Boltzmann machine (FVBM) and the other a RBM. A form of a reduction to a FVBM or an RBM comes into light if instead of modeling a transform error we relate just the linear transform or just the nonlinear transform representation ${\bf A}_{n-1}{\bf U}_{l-1}$ or ${\bf Y}_{l}$, respectively, under the backward propagation $\lambda_{l,f}=0$ or under the forward propagation $\lambda_{l,b}=0$.



\textbf{CNN}
Consider a reformulation of a convolution operation in a matrix vector form plus a composite goal that includes two subgoals. One related to a invariance over small translation, rotation and shift, and the other related to certain desired properties of the representation resulting from a nonlinear sub-sampling. In this case the goal is seen as a decomposition analysis operator resulting in a representation with a desired properties. The matrix vector convolution form is viewed as a single transform with a special transform matrix. In addition, the backward weight now will represent the deconvolution operator, again expressed in a matrix vector form. 

\textbf{R-NN} 
Consider an unfolded variant of the simplest, basic formulation of a R-NN \cite{Schmidhuber:2014:Overview}. Let only one transform matrix be defined, that is any two nodes are connected trough a same weight. It implies that the same weight is shared across all the time steps. The connection is evident if the hidden state is considered as a goal.

\textbf{AN} Although AN considers a generative modeling in a stochastic setup, the connection here is established w.r.t. a deterministic generator and discriminator functions. 
Under a deterministic functions 
there a connection exists to the concepts of \textit{active content fingerprint} and \textit{joint learning of linear feature map and content modulation} given in \cite{ Kostadinov:ICIP2016, Kostadinov:ICPR2016} and \cite{Kostadinov:EUSIPCO2017}. In these works the goal is to estimate two different functions under a min-max constraints/cost \cite{Kostadinov:ICIP2016}. Moreover, in their generalization the authors describes a min-max game 
that can be defined at input layer on the input data, at intermediate layer on the hidden representation or at the output layer on the output representation. Nevertheless they report results only for a single layer architecture with a min-max cost defined on the input layer. Instead, if we use the network equipped with the proposed modeling, two deterministic functions representing discriminator and generator, and a min-max cost then we have the connection to the AN. 

\textbf{VAE} The link between AN and VAE \cite{Kingma:2014:Adam} is established in \cite{YunchenPu:2017SVA_AN} under a symmetric costs and minimizing a variational lower bound. Where the adversarial solution comes as a natural consequence of symmetrizing the common VAE learning procedure. Moreover in that link the symmetric VAE cost is also seen as a log ratio test. Furthermore, under a deterministic mapping the log ratio test itself can be seen as a min-max cost. To arrive at the connection between VAE and the concept presented in this work  
we use the transform-based network, deterministic functions and a symmetric min-max cost.
The other link can be noticed by considering the learning principle of the VAE for the hole network, comprising of encoder and decoder part. Now if we use just two goals one as a constraint on the hidden representation for the estimation of the mean and the variance and the other for the reconstruction then we have the connection.


\bibliography{nips2018NLLPv10}

\begin{thebibliography}{46}
\providecommand{\natexlab}[1]{#1}
\providecommand{\url}[1]{\texttt{#1}}
\expandafter\ifx\csname urlstyle\endcsname\relax
  \providecommand{\doi}[1]{doi: #1}\else
  \providecommand{\doi}{doi: \begingroup \urlstyle{rm}\Url}\fi

\bibitem[Arjovsky et~al.(2017)Arjovsky, Chintala, and Bottou]{arjovsky:2017WAE}
M.~Arjovsky, S.~Chintala, and L.~Bottou.
\newblock {W}asserstein generative adversarial networks.
\newblock In D.~Precup and Y.~W. Teh, editors, \emph{Proceedings of the 34th
  International Conference on Machine Learning}, volume~70 of \emph{Proceedings
  of Machine Learning Research}, pages 214--223, International Convention
  Centre, Sydney, Australia, 06--11 Aug 2017. PMLR.

\bibitem[Baldi(2011)]{Baldi:2011AE}
P.~Baldi.
\newblock Autoencoders, unsupervised learning and deep architectures.
\newblock In \emph{Proceedings of the 2011 International Conference on
  Unsupervised and Transfer Learning Workshop - Volume 27}, UTLW'11, pages
  37--50. JMLR.org, 2011.
\newblock URL \url{http://dl.acm.org/citation.cfm?id=3045796.3045801}.

\bibitem[Balduzzi et~al.(2015)Balduzzi, Vanchinathan, and
  Buhmann]{Balduzzi:2015:KCB}
D.~Balduzzi, H.~Vanchinathan, and J.~Buhmann.
\newblock Kickback cuts backprop's red-tape: Biologically plausible credit
  assignment in neural networks.
\newblock In \emph{Proceedings of the Twenty-Ninth AAAI Conference on
  Artificial Intelligence}, AAAI'15, pages 485--491. AAAI Press, 2015.

\bibitem[Bengio(2012)]{Yoshua:2012:PR:GBT}
Y.~Bengio.
\newblock Practical recommendations for gradient-based training of deep
  architectures.
\newblock \emph{CoRR}, abs/1206.5533, 2012.

\bibitem[Bottou(2012)]{Bottou:2012:SGD_Tricks}
L.~Bottou.
\newblock Stochastic gradient descent tricks.
\newblock In \emph{Neural Networks: Tricks of the Trade - Second Edition},
  pages 421--436. 2012.

\bibitem[Cover and Thomas(2006)]{Cover:2006:EIT:1146355}
T.~M. Cover and J.~A. Thomas.
\newblock \emph{Elements of Information Theory (Wiley Series in
  Telecommunications and Signal Processing)}.
\newblock Wiley-Interscience, 2006.
\newblock ISBN 0471241954.

\bibitem[Czarnecki et~al.(2017)Czarnecki, Swirszcz, Jaderberg, Osindero,
  Vinyals, and Kavukcuoglu]{CzarneckiSJOVK:2017:Sintetic:Gradient}
W.~M. Czarnecki, G.~Swirszcz, M.~Jaderberg, S.~Osindero, O.~Vinyals, and
  K.~Kavukcuoglu.
\newblock Understanding synthetic gradients and decoupled neural interfaces.
\newblock \emph{CoRR}, abs/1703.00522, 2017.

\bibitem[Elad et~al.(2007)Elad, Milanfar, and
  Rubinstein]{elad07:analysisvsynthesis}
M.~Elad, P.~Milanfar, and R.~Rubinstein.
\newblock Analysis versus synthesis in signal priors.
\newblock \emph{Inverse Problems}, 23\penalty0 (3):\penalty0 947--968, June
  2007.

\bibitem[Gabriel(2017)]{Goh:2017:why_MW}
G.~Gabriel.
\newblock Why momentum really works.
\newblock \emph{Distill}, abs/7828, 2017.

\bibitem[Goodfellow et~al.(2014)Goodfellow, Pouget{-}Abadie, Mirza, Xu,
  Warde{-}Farley, Ozair, Courville, and Bengio]{Goodfellow:2014AN}
I.~J. Goodfellow, J.~Pouget{-}Abadie, M.~Mirza, B.~Xu, D.~Warde{-}Farley,
  S.~Ozair, A.~C. Courville, and Y.~Bengio.
\newblock Generative adversarial nets.
\newblock In \emph{Advances in Neural Information Processing Systems 27: Annual
  Conference on Neural Information Processing Systems 2014, December 8-13 2014,
  Montreal, Quebec, Canada}, pages 2672--2680, 2014.

\bibitem[He et~al.(2015)He, Zhang, Ren, and Sun]{HeZRS15}
K.~He, X.~Zhang, S.~Ren, and J.~Sun.
\newblock Deep residual learning for image recognition.
\newblock \emph{CoRR}, abs/1512.03385, 2015.

\bibitem[Hinton(2002)]{Hinton:2002:TPE}
G.~E. Hinton.
\newblock Training products of experts by minimizing contrastive divergence.
\newblock \emph{Neural Comput.}, 14\penalty0 (8):\penalty0 1771--1800, Aug.
  2002.

\bibitem[Hinton(2017)]{Hinton:2017BoltzmannMachines}
G.~E. Hinton.
\newblock Boltzmann machines.
\newblock In \emph{Encyclopedia of Machine Learning and Data Mining}, pages
  164--168. 2017.

\bibitem[Hochreiter(1998)]{Hochreiter:1998:VGP}
S.~Hochreiter.
\newblock The vanishing gradient problem during learning recurrent neural nets
  and problem solutions.
\newblock \emph{Int. J. Uncertain. Fuzziness Knowl.-Based Syst.}, 6\penalty0
  (2):\penalty0 107--116, Apr. 1998.

\bibitem[Jaderberg et~al.(2016)Jaderberg, Czarnecki, Osindero, Vinyals, Graves,
  and Kavukcuoglu]{JaderbergCOVGK:2016}
M.~Jaderberg, W.~M. Czarnecki, S.~Osindero, O.~Vinyals, A.~Graves, and
  K.~Kavukcuoglu.
\newblock Decoupled neural interfaces using synthetic gradients.
\newblock \emph{CoRR}, abs/1608.05343, 2016.

\bibitem[Kingma and Ba(2014)]{Kingma:2014:Adam}
D.~P. Kingma and J.~Ba.
\newblock Adam: {A} method for stochastic optimization.
\newblock \emph{CoRR}, abs/1412.6980, 2014.

\bibitem[Kingma and Welling(2013)]{KingmaW2013:VAE}
D.~P. Kingma and M.~Welling.
\newblock Auto-encoding variational bayes.
\newblock \emph{CoRR}, abs/1312.6114, 2013.

\bibitem[Kittel and Kroemer(1980)]{Kittel:Charles_Kroemer:Herbert_1980}
C.~Kittel and H.~Kroemer.
\newblock \emph{Thermal physics}.
\newblock W.H. Freeman, 2nd ed edition, 1980.

\bibitem[Kostadinov and Voloshynovskiy(2018)]{Kostadinov:ICLR2018}
D.~Kostadinov and S.~Voloshynovskiy.
\newblock Learning non-linear transform with discriminative and minimum
  information loss priors, 2018.
\newblock URL \url{https://openreview.net/pdf?id=SJzmJEq6W}.

\bibitem[Kostadinov et~al.(2016{\natexlab{a}})Kostadinov, Voloshynovskiy,
  Diephuis, Ferdowsi, and Holotyak]{Kostadinov:ICPR2016}
D.~Kostadinov, S.~Voloshynovskiy, M.~Diephuis, S.~Ferdowsi, and T.~Holotyak.
\newblock On local active content fingerprinting: solutions for general linear
  feature maps.
\newblock In \emph{ICPR}, Cancun, Mexico, December, 2-4 2016{\natexlab{a}}.

\bibitem[Kostadinov et~al.(2016{\natexlab{b}})Kostadinov, Voloshynovskiy,
  Diephuis, and Holotyak]{Kostadinov:ICIP2016}
D.~Kostadinov, S.~Voloshynovskiy, M.~Diephuis, and T.~Holotyak.
\newblock Local active content fingerprinting: optimal solution under linear
  modulation.
\newblock In \emph{ICIP}, Phoenix, USA, September, 25-28 2016{\natexlab{b}}.

\bibitem[Kostadinov et~al.(2017{\natexlab{a}})Kostadinov, Volshinovsky, and
  Ferdowsi]{Kostadinov:EUSIPCO2017}
D.~Kostadinov, S.~Volshinovsky, and S.~Ferdowsi.
\newblock Joint learning of local fingerprint and content modulation.
\newblock In \emph{EUSIPCO, Kos, Grece}, October, 1-5 2017{\natexlab{a}}.

\bibitem[Kostadinov et~al.(2017{\natexlab{b}})Kostadinov, Volshinovsky, and
  Ferdowsi]{SPARS:Kostadinov2017}
D.~Kostadinov, S.~Volshinovsky, and S.~Ferdowsi.
\newblock Learning non-structured, overcomplete and sparsifying transform.
\newblock In \emph{Lisbon, Portugal, SPARS 2017}, June 2017{\natexlab{b}}.

\bibitem[Lecun(1988)]{Lecun::TB:BP}
Y.~Lecun.
\newblock {A theoretical framework for Back-Propagation}.
\newblock 1988.

\bibitem[LeCun et~al.(1998)LeCun, Bottou, Orr, and M\"{u}ller]{LeCun:1998:EBT}
Y.~LeCun, L.~Bottou, G.~B. Orr, and K.-R. M\"{u}ller.
\newblock Efficient backprop.
\newblock In \emph{Neural Networks: Tricks of the Trade, This Book is an
  Outgrowth of a 1996 NIPS Workshop}, pages 9--50, London, UK, UK, 1998.
  Springer-Verlag.

\bibitem[LeCun et~al.(2015)LeCun, Bengio, and Hinton]{LeCunBH:2015:DL}
Y.~LeCun, Y.~Bengio, and G.~E. Hinton.
\newblock Deep learning.
\newblock \emph{Nature}, 521\penalty0 (7553):\penalty0 436--444, 2015.

\bibitem[Lee et~al.(2014)Lee, Zhang, Biard, and Bengio]{LeeZ:2014TP}
D.~Lee, S.~Zhang, A.~Biard, and Y.~Bengio.
\newblock Target propagation.
\newblock \emph{CoRR}, abs/1412.7525, 2014.

\bibitem[Lipton(2015)]{Lipton:2015RNN}
Z.~C. Lipton.
\newblock A critical review of recurrent neural networks for sequence learning.
\newblock \emph{CoRR}, abs/1506.00019, 2015.

\bibitem[Loshchilov and Hutter(2016)]{Loshchilov:2016:SGDR}
I.~Loshchilov and F.~Hutter.
\newblock {SGDR:} stochastic gradient descent with restarts.
\newblock \emph{CoRR}, abs/1608.03983, 2016.

\bibitem[Makhzani et~al.(2015)Makhzani, Shlens, Jaitly, and
  Goodfellow]{Makhzani:2015AAE}
A.~Makhzani, J.~Shlens, N.~Jaitly, and I.~J. Goodfellow.
\newblock Adversarial autoencoders.
\newblock \emph{CoRR}, abs/1511.05644, 2015.

\bibitem[N\o~kland(2016)]{Nokland:2016DF}
A.~N\o~kland.
\newblock Direct feedback alignment provides learning in deep neural networks.
\newblock In D.~D. Lee, M.~Sugiyama, U.~V. Luxburg, I.~Guyon, and R.~Garnett,
  editors, \emph{Advances in Neural Information Processing Systems 29}, pages
  1037--1045. Curran Associates, Inc., 2016.

\bibitem[Pascanu et~al.(2012)Pascanu, Mikolov, and
  Bengio]{Razvan:Pascanu:Exp:Grad:prob}
R.~Pascanu, T.~Mikolov, and Y.~Bengio.
\newblock Understanding the exploding gradient problem.
\newblock \emph{CoRR}, 2012.

\bibitem[Plaut et~al.(1986)Plaut, Nowlan, and Hinton]{Plaut:1986:BP}
D.~C. Plaut, S.~J. Nowlan, and G.~E. Hinton.
\newblock Experiments on learning back propagation.
\newblock Technical Report CMU--CS--86--126, Carnegie--Mellon University,
  Pittsburgh, PA, 1986.

\bibitem[Pu et~al.(2017)Pu, Chen, Dai, Wang, Li, and
  Carin]{YunchenPu:2017SVA_AN}
Y.~Pu, L.~Chen, S.~Dai, W.~Wang, C.~Li, and L.~Carin.
\newblock Symmetric variational autoencoder and connections to adversarial
  learning.
\newblock \emph{CoRR}, abs/1709.01846, 2017.

\bibitem[Ravishankar and Bresler(2014)]{DBLP:conf/icassp/RavishankarB14}
S.~Ravishankar and Y.~Bresler.
\newblock Doubly sparse transform learning with convergence guarantees.
\newblock In \emph{{IEEE}, {ICASSP} 2014, Florence, Italy, May 4-9, 2014},
  pages 5262--5266, 2014.

\bibitem[Rubinstein and Elad(2014)]{RubinsteinE14}
R.~Rubinstein and M.~Elad.
\newblock Dictionary learning for analysis-synthesis thresholding.
\newblock \emph{{IEEE} Trans. Signal Processing}, 62\penalty0 (22):\penalty0
  5962--5972, 2014.

\bibitem[Ruder(2016)]{Ruder:2016:Overview}
S.~Ruder.
\newblock An overview of gradient descent optimization algorithms.
\newblock \emph{CoRR}, abs/1609.04747, 2016.

\bibitem[Schmidhuber(2014)]{Schmidhuber:2014:Overview}
J.~Schmidhuber.
\newblock Deep learning in neural networks: An overview.
\newblock \emph{CoRR}, abs/1404.7828, 2014.

\bibitem[Shamir and Zhang(2013)]{Shamir:2013:SGD_NON_Smood}
O.~Shamir and T.~Zhang.
\newblock Stochastic gradient descent for non-smooth optimization: Convergence
  results and optimal averaging schemes.
\newblock In \emph{Proceedings of the 30th International Conference on Machine
  Learning, {ICML} 2013, Atlanta, GA, USA, 16-21 June 2013}, pages 71--79,
  2013.

\bibitem[Spivak(1980)]{spivak1980calculus}
M.~Spivak.
\newblock \emph{Calculus}.
\newblock Addison-Wesley world student series. Publish or Perish, 1980.
\newblock URL \url{https://books.google.ch/books?id=-mwPAQAAMAAJ}.

\bibitem[Srivastava et~al.(2014)Srivastava, Hinton, Krizhevsky, Sutskever, and
  Salakhutdinov]{Srivastava:2014:Dropout}
N.~Srivastava, G.~Hinton, A.~Krizhevsky, I.~Sutskever, and R.~Salakhutdinov.
\newblock Dropout: A simple way to prevent neural networks from overfitting.
\newblock \emph{J. Mach. Learn. Res.}, 15\penalty0 (1):\penalty0 1929--1958,
  Jan. 2014.

\bibitem[Taylor et~al.(2016)Taylor, Burmeister, Xu, Singh, Patel, and
  Goldstein]{Taylor:2016BXSPG}
G.~Taylor, R.~Burmeister, Z.~Xu, B.~Singh, A.~Patel, and T.~Goldstein.
\newblock Training neural networks without gradients: {A} scalable {ADMM}
  approach.
\newblock \emph{CoRR}, abs/1605.02026, 2016.

\bibitem[Vapnik(1995)]{Vapnik:1995:SL:ER}
V.~N. Vapnik.
\newblock \emph{The Nature of Statistical Learning Theory}.
\newblock Springer-Verlag New York, Inc., New York, NY, USA, 1995.
\newblock ISBN 0-387-94559-8.

\bibitem[Vincent et~al.(2010)Vincent, Larochelle, Lajoie, Bengio, and
  Manzagol]{VincentLLBM10}
P.~Vincent, H.~Larochelle, I.~Lajoie, Y.~Bengio, and P.~Manzagol.
\newblock Stacked denoising autoencoders: Learning useful representations in a
  deep network with a local denoising criterion.
\newblock \emph{Journal of Machine Learning Research}, 11:\penalty0 3371--3408,
  2010.

\bibitem[Wan et~al.(2013)Wan, Zeiler, Zhang, Cun, and Fergus]{pmlr-v28-wan13}
L.~Wan, M.~Zeiler, S.~Zhang, Y.~L. Cun, and R.~Fergus.
\newblock Regularization of neural networks using dropconnect.
\newblock In S.~Dasgupta and D.~McAllester, editors, \emph{Proceedings of the
  30th International Conference on Machine Learning}, volume~28 of
  \emph{Proceedings of Machine Learning Research}, pages 1058--1066, Atlanta,
  Georgia, USA, 17--19 Jun 2013. PMLR.

\bibitem[Zhu et~al.(2017)Zhu, Meng, and Zhang]{Zhu:Adam_Look_Ahead}
A.~Zhu, Y.~Meng, and C.~Zhang.
\newblock An improved adam algorithm using look-ahead.
\newblock In \emph{Proceedings of the 2017 International Conference on Deep
  Learning Technologies}, ICDLT '17, pages 19--22, New York, NY, USA, 2017.
  ACM.

\end{thebibliography}
\bibliographystyle{abbrvnat}




\clearpage
\newpage

\end{document}